\documentclass{article}

\usepackage[dvipsnames,table]{xcolor} 
\usepackage[preprint]{neurips_2025}

\usepackage[utf8]{inputenc} 
\usepackage[T1]{fontenc}    
\usepackage{hyperref}       
\usepackage{url}            
\usepackage{booktabs}       
\usepackage{amsfonts}       
\usepackage{nicefrac}       
\usepackage{microtype}      
 \usepackage{enumitem}

\usepackage{amsmath}
\usepackage{graphicx}

\title{Advertising in AI systems:
Society must be vigilant}

\author{%
  Menghua Wu \\
  Department of Computer Science\\
  Massachusetts Institute of Technology\\
  \texttt{rmwu\{at\}mit.edu} \\
  \And
  Yujia Bao \\
  Center for Advanced AI \\
  Accenture \\
  \texttt{yujia.bao\{at\}accenture.com} \\
}

\begin{document}

\maketitle

\begin{abstract}
AI systems have increasingly become our gateways to the Internet.
We argue that just as advertising has driven the monetization of web search and social media, so too will commercial incentives shape the content served by AI.
Unlike traditional media, however, the outputs of these systems are dynamic, personalized, and lack clear provenance -- raising concerns for transparency and regulation.
In this paper, we envision how commercial content could be delivered through generative AI-based systems.
Based on the requirements of key stakeholders -- advertisers, consumers, and platforms -- we propose design principles for commercially-influenced AI systems.
We then outline high-level strategies for end users to identify and mitigate commercial biases from model outputs.
Finally, we conclude with open questions and a call to action towards these goals.
\end{abstract}
\section{Introduction}

Over the past several years, AI systems have become ubiquitous among the general public, partly due to their ability to distill online knowledge into personalized recommendations~\citep{ouyang2022training}.
Their explosive growth mirrors that of search engines~\citep{brin1998anatomy} and social media~\citep{wilson2012review} in the early 2000s -- both of which introduced new paradigms for consuming media online.
Following market dominance, these platforms adopted business models driven by the monetization of user data, often through advertising~\citep{kenney2016rise}.
Firms that offer goods or services aim to identify users who are likely to purchase, and digital platforms that collect personal data are well-positioned to make the connection~\citep{bagwell2007economic,evans2008economics,lau2020brief}.
Advertising is a multi-billion dollar market, with global spending on search engines and social media projected to exceed \$600 billion in 2025.\footnote{\url{https://www.statista.com/outlook/amo/advertising/worldwide}}
As the reach of AI systems continues to expand, it is inevitable that these commercial forces will reshape AI-mediated content delivery.
To ensure that AI systems remain transparent and fair, \textbf{the machine learning community must take an active stance in developing standards, auditing practices, and algorithmic safeguards that anticipate and mitigate commercial exploitation.
}

The differences between traditional digital platforms and generative AI systems complicate the serving and regulation of commercial content.
Over the past decade, digital platforms have become increasingly skilled at collecting personal information for marketing purposes~\citep{lau2020brief}.
However, the nature of the advertisement has remained virtually unchanged:
the same piece of static, commercial content is shown to large groups of users~\citep{williamson1978decoding}.
Since advertisements have always contained predetermined information, albeit presented through evolving media, the core principles behind their regulation have remained consistent since the late twentieth century~\citep{pitofsky1976beyond,peltzman1981effects}.
In a key departure from classical advertising, \textbf{generative AI systems have the capacity to \emph{dynamically} produce commercial content, designed to explicitly or surreptitiously monetize each and every interaction}.
These systems are able to interact with users at unprecedented levels of specificity and personalization~\citep{ouyang2022training,rombach2022high}.
Unlike search engines or social media, which offer coarse sets of results, generative AI systems deliver tailored responses that address each user's exact needs.
This capacity for personalization creates an environment ripe for commercial exploitation.
As AI systems become indispensable tools for information seeking and decision-making, the incentives to influence their outputs grow commensurately.
Without proper safeguards, commercially-motivated content -- advertisements, product placements, or strategically presented information -- may become indistinguishable from objective information.

This position paper examines how commercial incentives might shape AI systems in the years to come, with explicit parallels to the monetization of web search and social media.
Our analysis is centered around the notion of AI as an advertising platform, whose goal is to connect advertisers to appropriate consumers.
Based on this framework, we introduce settings in which commercial content might manifest in the outputs of generative AI models (Section~\ref{sec:incorporate}), and describe how these AI systems may be built to accommodate such content (Section~\ref{sec:system}).
Finally, we survey the open problems that emerge from such a system (Sections~\ref{sec:bias}-\ref{sec:open-questions}) -- including how users might identify and eliminate commercial biases; and how AI platforms might train on data that has been contaminated with advertisements.

We believe that there are three pillars to the responsible commercialization of AI systems in the context of digital marketing.
\begin{enumerate}
\item Governments must regulate advertising and other forms of commercial influence in AI systems, with guidelines for consumer protection.
\item AI platforms must ensure that commercial incentives are transparent, and that the inclusion of sponsored content does not lead to harmful outputs or degraded models over time.
\item Researchers must develop algorithmic guardrails and benchmarks that assess whether commercialized AI platforms are safe for consumers and adherent to regulations.
\end{enumerate}
In this work, we primarily focus on the second and third axes, with recommendations for industry and the research community.



\section{How might commercial incentives be incorporated into AI systems?}
\label{sec:incorporate}

\begin{figure}
\centering
\includegraphics[width=\linewidth]{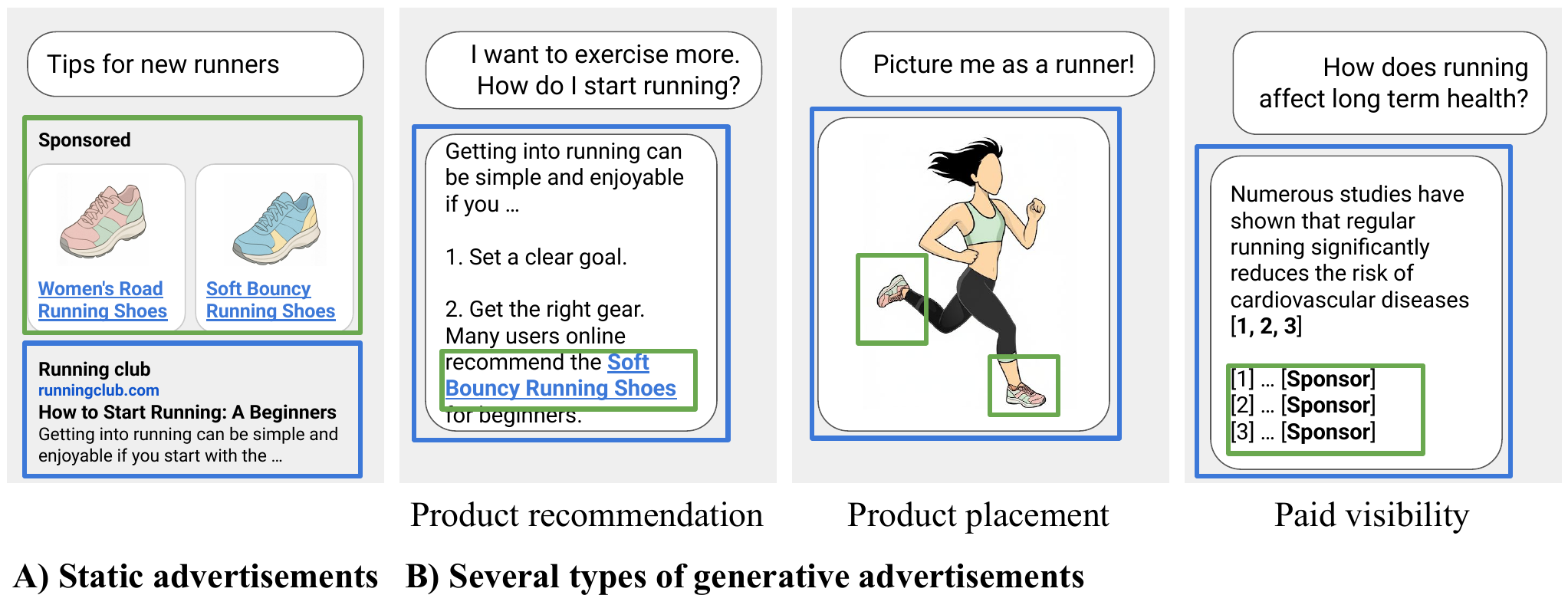}
\caption{(A) Classic advertisements are independent of the platform's primary media. Sponsored content (green) does \emph{not} impact ordinary content (blue), and they can be inserted anywhere.
(B) Generative advertisements are AI-generated responses which have been conditioned on both the user query and sponsored information. They \emph{do} affect the primary outputs of AI systems.
}
\vspace{-0.2in}
\label{fig:definition}
\end{figure}

There are two types of sponsored content that may be served alongside AI systems (Figure~\ref{fig:definition}).
We use the term \emph{static advertisements} to denote advertisements whose content appears the same to every viewer, and are primarily independent pieces of media~\citep{williamson1978decoding}.
For example, a video-sharing platform may require that users watch a short advertisement before providing access to the actual content that they clicked on~\citep{firat2019youtube}.
Other forms may include sponsored search results~\citep{lee2011google}, social media posts~\citep{dehghani2015research}, or news articles~\citep{carlson2015news}.
While static advertisements are most effective when served to an appropriate audience, in principle they can be inserted anywhere -- AI systems included.
In fact, today's AI systems already promote premium subscription plans in prominent locations (Figure~\ref{fig:ad}).
Barring aesthetic factors, it is not hard to imagine the inclusion of third-party advertisements elsewhere.

Generative AI introduces an additional dimension.
Suppose a user inputs query $x$.
An AI system draws upon pieces of information $z := \{z_i\}_{i=1}^n$ to generate output $y$.
In current frameworks, $z_i$ may include the model's innate knowledge, relevant documents~\citep{lewis2020retrieval}, or the results of tool use~\citep{zhuang2023toolqa}.
However, $z$ may also contain sponsored content that would not otherwise have been retrieved.
For example, if a user asks about the best running shoes under \$100, a new shoe company could pay for their product to be discussed among established brands.
We use \emph{generative advertisements} to denote outputs $y$ which were generated based on $z$ that contains commercial incentives~\citep{Feizi2023OnlineAW}.
Among the broader literature, advertisements which look identical to their surrounding media are also known as \emph{native advertisements}~\citep{carlson2015news,campbell2019challenges},
which encompass our definition of generative advertisements.

To provide concrete examples of generative advertisements, which are the primary focus of this work, Section~\ref{sec:types} introduces several types of human-AI interactions that are primed for monetization.
Next, Section~\ref{sec:pro-con} discusses comparative advantages and challenges of these new modalities, compared to static advertisements.
Finally, we note that generative advertisements are fundamentally different from using generative AI to automate the creation of static advertisements~\citep{huh2023chatgpt,malikireddy2024generative,osadchaya2024chatgpt}.
The key distinction is that AI-generated static advertisements are created offline, rather than online during an interaction with a specific user.

\subsection{Opportunities for generative advertising}
\label{sec:types}

\paragraph{Product recommendation}

Users may directly solicit recommendations for products or services~\citep{Chang2024ACS}.
Since the expected response is a list of actionable purchases, it would be natural to include one or more sponsored listings~\citep{Feizi2023OnlineAW}.
There are also user queries that do not explicitly express an intent to purchase, but product recommendations would not seem out of place.
For example, it is not uncommon for patients and clinicians to solicit medical advice from AI systems~\citep{hosseini2023exploratory,ghassemi2023chatgpt}.
Given a set of symptoms as input, it would be quite reasonable to respond with a list of remedies.
In these scenarios, the boundary between genuine assistance and advertising is unclear, leading to opportunities for unsolicited, stealth marketing.

\paragraph{Product placement}

Advertising may also take the form of product placement, in which a product or service is featured in the background of other media~\citep{williams2011product}.
A classic example is when actors to use or refer to branded products in movies~\citep{gupta1998product}.
Product placement is widely applicable to generative AI systems, where commercial products may be promoted in the place of generic alternatives.
In the context of language models, a coding assistant may be told to use commercial software instead of open source alternatives.
Multimodal generative AI~\citep{rombach2022high} may also be prone to such influences.
For example, if a user asked an AI system to draw their portrait as a celebrity, it would be natural to depict recognizable luxury items.
Likewise, if an interior designer were to re-imagine a space through generative AI~\citep{Wang2024RoomDreamingGA}, they could become unsuspecting targets of product placement by a furniture design firm.
In all three cases, if the product placements were not explicitly disclosed, they could be difficult to detect -- leading to concerns about transparency.


\paragraph{Paid visibility}

Commercial influences may cause AI systems to feature information from certain knowledge sources over others.
For example, scientific publishers might pay for their portfolios to be favored by AI research assistants~\citep{skarlinski2024language}, resulting in increased citations and readership over time.
AI platforms could also charge fees for accessing these licensed materials, leading to new streams of revenue for publishers and platforms.
These commercial relationships are not inherently harmful and could increase the credibility of research assistants.
However, they also have the potential to exacerbate biases in the production of and access to scientific literature~\citep{salager2008scientific,hausmann2016challenges}.
Furthermore, an infrastructure that allows for preferential treatment of knowledge can also be used to push a political agenda or otherwise shape public opinion~\citep{Rozado2023ThePB,Fisher2024BiasedAC}.
For example, a recent study found that social media platform X's paid verification system actually amplified low credibility content, compared to when paid options for verification were not available~\citep{corsi2024evaluating}.
Thus, platforms must take extreme care with respect to the monetization of information visibility.



\subsection{Advantages and challenges of generative advertising}
\label{sec:pro-con}

Generative AI makes it possible to tailor commercial content to individual interactions.
This flexibility is also a major challenge in the implementation and governance of generative advertisements.

\paragraph{Specificity and suitability} Users interact with AI systems at much higher resolution than with search engines or social media, where personal data may ``only'' consist of search history~\citep{matthijs2011personalizing} and relevance feedback~\citep{broder2008search}.
Thus, it is natural for AI recommendations to be more fine-grained than current options.
In contrast to static advertisements, however, there is a tacit expectation that generative AI outputs should directly address user queries~\citep{skjuve2024people}.
Commercial material must be integrated into model responses in ways that feel coherent to the user.
Thus, the increased capacity for customization is accompanied by a stricter requirement for assessing content relevance, and a higher bar for adapting commercial content to each situation.



\paragraph{Long term impact on platform quality}

The independence of static advertisements from a digital platform's primary media is both a limitation and a strength.
While users could become irritable in response to poorly selected ads~\citep{firat2019youtube}, static advertisements do not otherwise degrade the user experience on each platform, whose media is produced by separate content creators~\citep{gerhards2019product}.
In the context of generative advertisements, however, the inclusion of sponsored media may adversely affect model quality in the long term, due to the close integration of commercial and non-commercial content.
Specifically, human feedback on model generations is crucial for the alignment of AI systems with desired traits like usefulness, factuality, or safety~\citep{ouyang2022training,wang2023aligning,ji2023beavertails}.
If commercial content is naively included as part of these alignment datasets, AI models may inadvertently ``forget'' that these content are sponsored -- resulting in sustained commercial biases over time.
On the other hand, if these outputs are entirely stripped from the data, user responses may appear out of context -- teaching models to hallucinate about information that (apparently) does not exist.
Thus, proper processing of commercially-influenced human feedback is crucial for ensuring the longevity of these systems.

\paragraph{Ease of regulation}

While there are some concerns about how advertising is regulated in practice, especially with respect to native advertisements~\citep{carlson2015news,campbell2019challenges}, it is generally clear how one would verify whether commercial content is compliant~\citep{pitofsky1976beyond,peltzman1981effects}.
Digital advertising must be disclosed in a clear and conspicuous manner~\citep{Evans2017DisclosingII}; health claims must be backed by solid proof~\citep{Villanueva2003AccuracyOP,Williams2005ConsumerUA}; and so on.
In contrast to traditional media, the outputs of generative AI are stochastic, and they vary immensely based on individual input prompts~\citep{wei2022chain} or decoding strategies~\citep{wangchain}.
As a result, it may be difficult to reproduce, or even verify incompliant model behavior.

\section{How do we design an AI system that serves commercial content?}
\label{sec:system}

Understanding how AI systems might serve commercial content is a prerequisite to their regulation and monitoring.
In this section, we first introduce the stakeholders and describe their expectations towards an AI system that supports advertising (Section~\ref{sec:stakeholders}).
Based on these requirements and properties of current advertising systems, we delineate the systems design principles that must be fulfilled (Section~\ref{sec:system-design}).
Finally, we describe how such a system may be implemented in the context of large language models (LLMs) and multi-agent systems (Section~\ref{sec:implementation}).

\subsection{Who are the stakeholders, and what are their requirements?}
\label{sec:stakeholders}

\begin{figure}
\centering
\includegraphics[width=\linewidth]{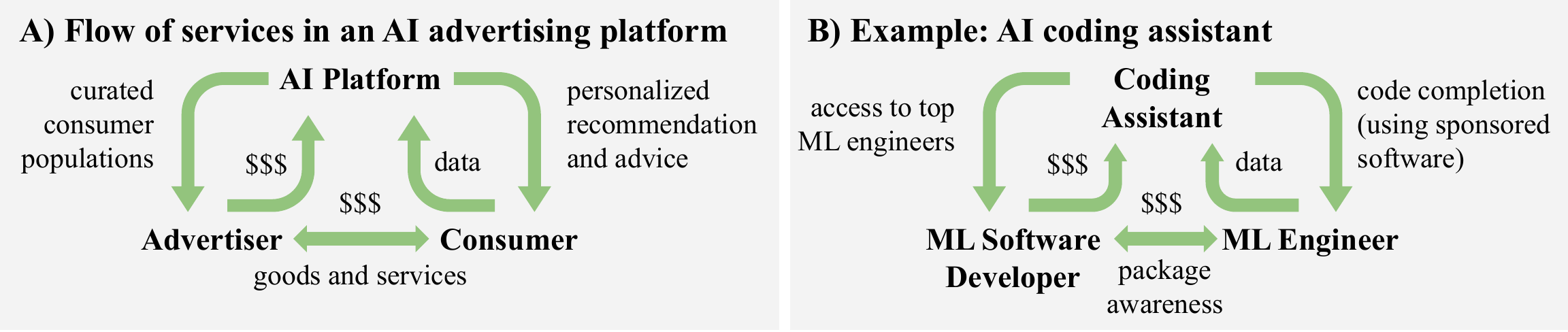}
\caption{Flow of services in an AI system. A) Abstract overview. B) A coding assistant may serve as a platform that connects software vendors to software consumers (engineers).}
\vspace{-0.2in}
\label{fig:cycle}
\end{figure}

There are three key players in the digital advertising economy: the advertiser, the consumer, and the platform~\citep{kenney2016rise,lau2020brief}.
On a high level, the platform creates value by connecting firms, who wish to sell, to consumers, who wish to buy (Figure~\ref{fig:cycle}).

\paragraph{Advertiser}

Firms that offer good or services spend on advertising to disseminate information about their products to customers who may be interested in purchasing.
In exchange, they expect to receive benefits like increased sales, brand loyalty, or other competitive advantages~\citep{bagwell2007economic}.
Since these long term goals are difficult to price, there are two models by which digital advertises are billed: pay-per-click~\citep{kritzinger2013search} and pay-per-view~\citep{mangani2004online}.
This arrangement rests upon the trust that \textbf{the platform must serve the commercial content, whenever it claims to have done so}.
Failure to provide accurate statistics about impressions, clicks, or conversions will erode confidence and undermine the platform itself.
While obvious, this level of fidelity may or may not be easy to implement, as the outputs of generative AI models are non-deterministic.

\paragraph{Consumer}

Users interact with AI systems for a variety of reasons, including research, education, entertainment, and healthcare ~\citep{choudhury2023investigating,hosseini2023exploratory}.
Regardless of intent, consumers will only use a digital platform as long as it fulfills basic user needs, such as providing correct answers or relevant information~\citep{skjuve2024people}.
Thus, \textbf{the incorporation of commercial content must not compromise the system's normal operation}.
Furthermore, if an AI system chooses to monetize user interactions, \textbf{consumers should be able to opt out of targeted advertising and other forms of commercial influence}.
There are several motivations for platforms to implement this concept.
It may be required by law~\citep{goldfarb2011privacy};
this option may be monetized itself~\citep{anderson2009free,bapna2018monetizing}; or it may allow firms to focus on users who are more receptive and less sensitive to targeted advertising~\citep{johnson2020consumer}.


\paragraph{Platform}

Significant burden of implementation falls upon the platform, which facilitates the relationship between advertisers and consumers.
Beyond its obligations to the other two parties, the platform must ensure its commercial viability and regulatory compliance.

Pricing is central to the economics of advertising.
In current digital platforms, the amount paid for each click or view is determined by auction, in which advertisers bid to serve their ads to specific users~\citep{borgs2007dynamics,yuan2013real,yuan2014survey}.
To support dynamic pricing structures and an open market, \textbf{AI platforms must be able to insert commercial recommendations in real time, per interaction}.
During digital auctions, there are two primary factors that contribute to pricing algorithms: the relevance of each ad to each user, and each user's intent to purchase~\citep{solberg2024using}.
Users who are more likely to relate with a certain ad cost less, and users who are more likely to complete a transaction cost more.
To implement such algorithms, \textbf{AI platforms must efficiently quantify relevance and user intent}.

There are also legal guidelines that dictate how ads must be disclosed, and what content can or must be included in them.\footnote{\url{https://www.ftc.gov/business-guidance/advertising-marketing}}
To ensure that their advertisements are compliant, \textbf{AI platforms must maintain provenance of each aspect of their generated content}.
Establishing direct mappings between information sources and model outputs will permit appropriate disclosures and content verification.








\begin{figure}
\centering
\includegraphics[width=\linewidth]{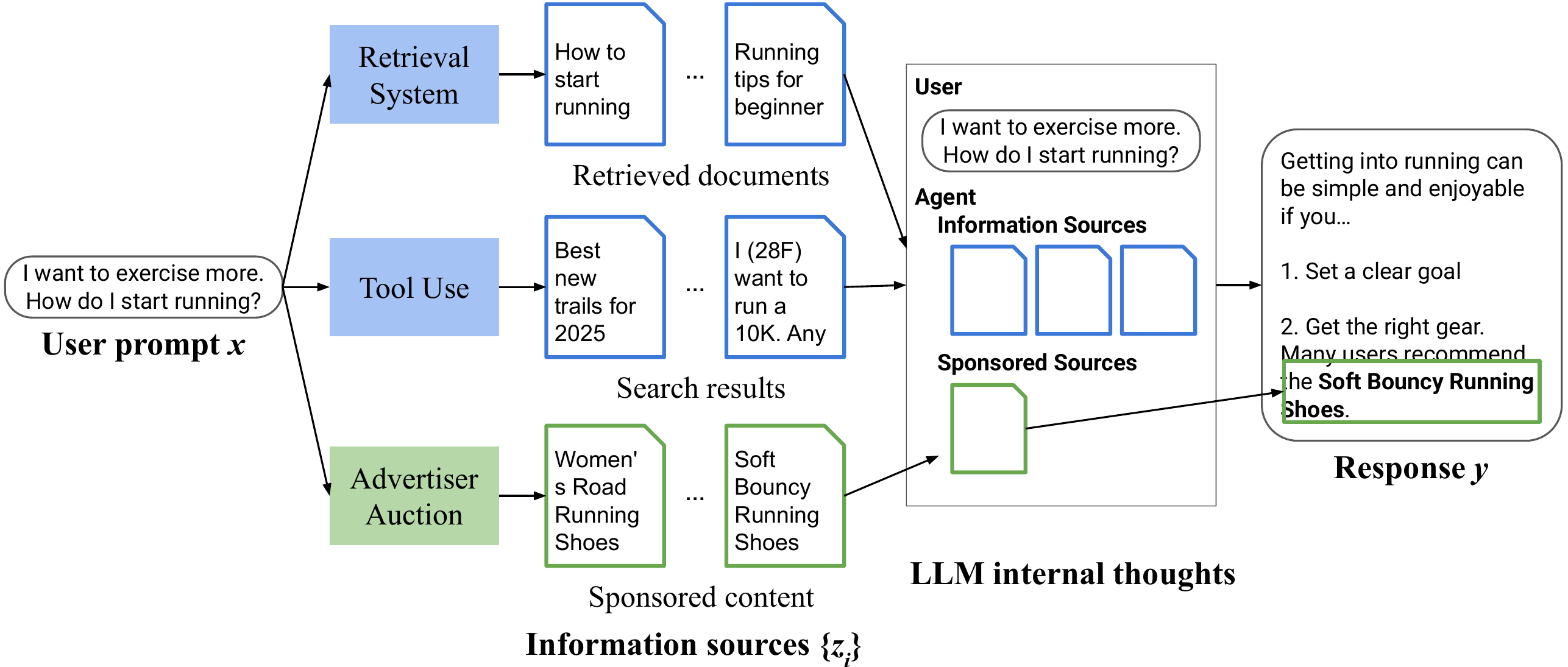}
\caption{
Example of AI system with advertisements.
Generative advertisements derive their content from both traditional information sources, as well as additional sponsored content.
These sources may be accessed via standard tool-call patterns and synthesized by a language model into an overall response.
}
\label{fig:system}
\vspace{-0.2in}
\end{figure}






\subsection{System design principles}
\label{sec:system-design}
We introduce four core principles that any AI system with sponsored content must fulfill. Let $x$ denote the user prompt, and let $z = \{z_1,\dots,z_n\}$ represent all information retrieved from external sources (including sponsor-provided text). We partition $z$ into $z_{\text{w/o ad}}$ (non-sponsored) and $z_{\text{ad}}$ (advertisements). The final model output may exclude ads, $y^{\text{w/o ad}}(x)$, or include them, $y^{\text{w/ad}}(x)$.

\paragraph{Faithfulness}
Generative advertisements must remain factually accurate, with respect to the sponsored source material.
Define a fidelity metric
\[
F\bigl(y^{\text{w/ad}}(x), z_i\bigr), \quad z_i \in z_{\text{ad}},
\]
which is maximized when $z_i$ is perfectly captured by the final output $y^{\text{w/ad}}(x)$.
The system should ensure $F(\cdot) \ge \delta$ for a threshold $\delta>0$, thus preserving advertiser trust and maintaining regulatory compliance.
In practice, $F$ could denote logical entailment from $z_i$ to $y$~\citep{maccartney2014natural,tatsu2024factuality} or other forms of factuality~\citep{weilong}.

\paragraph{Utility}
Commercial insertions must not degrade the system’s core functions. Let $U\bigl(y^{\text{w/ad}}(x)\bigr)$ and $U\bigl(y^{\text{w/o ad}}(x)\bigr)$ be utility measures for outputs with and without ads, respectively. We require:
\[
U\bigl(y^{\text{w/ad}}(x)\bigr)\;\ge\;\alpha \cdot U\bigl(y^{\text{w/o ad}}(x)\bigr),
\]
for some $\alpha \in (0,1]$, ensuring acceptable quality even when ads are present.
The definition of $U$ is flexible and may encompass multiple objectives, including including relevance, latency, and accuracy~\citep{wang2024understanding}.

\paragraph{Privacy}
Each user may select $o_u \in \{\textsc{in}, \textsc{out}\}$ to indicate whether personal data $\mathcal{D}(u)$ may be used for ad selection. If $o_u=\textsc{out}$, ad retrieval must proceed without accessing $\mathcal{D}(u)$, thereby delivering generic (non-personalized) ads.

\paragraph{Provenance}
The system must track the origin of tokens in $y^{\text{w/ad}}(x)$ to ensure compliance and transparency. Define a mapping
\[
\phi : y_i \to (0, 1)^{\{z_1,\dots,z_n\}},
\]
where each output token $y_i$ is associated with the subset of $z$ that contributed to it. Maintaining this mapping supports regulatory audits and internal monitoring of sponsor content.

\subsection{Implementing the AI system}
\label{sec:implementation}
We present an approach that upholds the above principles by leveraging standard ``tool-call'' patterns in large language models~\citep{team2023gemini,grattafiori2024llama,hurst2024gpt}.
In this setup, tool calls occur whenever external information or additional computation is needed, and acquiring advertisements is treated as one such tool call that runs concurrently upon receipt of the user’s query.
\begin{enumerate}
    \item \textbf{User prompt:}  
    The user provides a prompt $x$.  
    
    \item \textbf{Concurrent tool calls:}  
    The LLM initiates parallel tool calls. For example:  
    \begin{itemize}
        \item The \textbf{content tool call} gathers information relevant to $x$, using tools like search, retrieval-augmented generation~\citep{lewis2020retrieval}, or other external APIs. This yields $z_{\text{w/o ad}}$.
        \item The \textbf{advertisement tool call} acquires sponsor content conditioned on $x$ and the user’s preference $o_u$, resulting in $z_{\text{ad}}$.
        This step may also involve a pricing auction among the potential advertisers~\citep{yuan2013real,duetting2024mechanism}.
    \end{itemize}
    
    \item \textbf{(Optional) Chain-of-thought:}  
    The LLM processes the non-sponsored content, generating an internal reasoning trace to refine the context and consolidate information~\cite{guo2025deepseek}.
    This is not always necessary for sufficiently simple questions.
    \item \textbf{Ad insertion in the LLM context:}  
    The system may append commercial content to the model’s input with a special tag, such as
    \[
      \texttt{<ad>} \; z_{\text{ad}} \; (\text{sponsored content}) \; \texttt{</ad>}.
    \]
    This marking makes explicit which tokens are drawn from sponsored text, whose information must be closely preserved.
    \item \textbf{Final generation:}  
    The LLM produces a response $y^\text{w/ad}$ that integrates both non-sponsored and sponsor-provided content, subject to any required disclaimers or user opt-outs.
\end{enumerate}
\paragraph{Lightweight post-generation analysis and model state reuse}
Since both the final output $y^{\text{w/ad}}$ and the ad materials $z_{\text{ad}}$ reside within the same context, modern LLMs can easily perform follow-up analysis. After generating $y^{\text{w/ad}}$, the model can reuse its key-value (KV) cache and simply continue its generation to produce.
For example, we may append the section
\[
  \texttt{<ad-report>} \; \ldots \; \texttt{</ad-report>},
\]
which verifies how faithfully $y^{\text{w/ad}}$ represents $z_{\text{ad}}$, and whether sponsor requirements were satisfied. By leveraging the cached states, the system saves computational overhead for this verification step.

\paragraph{Training for advertisement placement}
Model providers can employ standard training strategies to improve how LLMs integrate advertisements.

\begin{itemize}
    \item \textbf{Supervised fine-tuning:} The model developer provides ``golden'' examples of well-placed ads, which adhere to best practices for factuality, disclaimers, and user experience.
    By learning from these curated examples, the model becomes more proficient at blending ads into user-facing content~\citep{ouyang2022training,chung2024scaling,liu2023visual}.
    \item \textbf{Implicit rewards based on user interactions:}
    After an ad is displayed, the user’s response (e.g., acceptance, neutrality, or rejection) can be classified via another short LLM call that reuses the existing context (and thus the KV cache). This feedback signal can serve as a reward function for further alignment~\citep{rafailov2023direct,kumar2024training,guo2025deepseek}.
\end{itemize}

\section{Identifying and mitigating commercial biases}
\label{sec:bias}

Embedding advertisements into an AI system response can cause the model to favor certain brands or products at the expense of user experience or model quality.
Furthermore, since the regulatory environment around generative advertisements is in its infancy, we must develop algorithmic safeguards for consumer protection and model surveillance.
Unlike traditional concerns around
factual accuracy and misinformation~\citep{wang2017liar,Thorne18Fever,guo2022survey,tatsu2024factuality},
commercial content does not necessarily introduce falsehoods.
Rather, they subtly familiarize users with a sponsor’s offerings.
Thus, we seek an algorithm
that can identify and neutralize promotional content, such that users may receive an answer unadulterated by
commercial influences.

These methods may draw inspiration from the bias and fairness literature.
Current debiasing techniques for large language models can be broadly categorized
as \emph{training-based} or \emph{inference-based} approaches. Training-stage
methods~\citep{webster2020measuring,ghanbarzadeh2023gender,limisiewicz2023debiasing}
use data augmentation or explicit regularization to reduce social or demographic
biases, but they can be computationally intensive and difficult to scale for
billion-parameter models. Inference-stage methods avoid retraining and rely on
techniques such as debiasing prompts~\citep{hida2024social,si2022prompting},
projection-based corrections~\citep{ravfogel2020null}, or explicit neuron
suppression~\citep{dasu2024neufair,xiao2025fairness}. However, they can struggle
to comprehensively remove biases of all forms. Here, we focus on
\emph{commercial bias}, which differs from social or ideological bias in that these biases are not innate to a model's weights; its distribution may shift dynamically; and the promotional content may be low quality or factually incorrect.


\paragraph{Debiasing in the context of advertisements}
We define \emph{debiasing} as the process of recovering, from a final
ad-augmented output \(y^{\text{w/ad}}(x)\), the original or near-original model
response \(y^{\text{w/o ad}}(x)\) that the system would have produced
\emph{without} exposure to sponsor content.
That is, we aim to remove brand
references, promotional phrasing, and other marketing language while preserving
the response's correctness and central user intent.
This allows users to ``block'' commercial content, similar to how advertisements may be hidden in search engines~\citep{iqbal2017ad}.

We discuss two different methods to generate an ad-free version of the
LLM’s response: (1) direct debiasing (a single-pass transformation with a
specialized model), and (2) multi-sampling and aggregation.

\subsection{Approach 1: Direct debiasing (single-pass transformation)}

The first approach is to process the ad-influenced output \(y^{\text{w/ad}}(x)\) through a specialized, ad-free ``debiasing'' model.

\begin{enumerate}
    \item \textbf{Debiasing procedure:}
    \(y^{\text{w/ad}}(x)\) is processed through a separate LLM, configured to identify and remove
    commercial content. The user can
    also include preferences in the prompt (e.g. avoid brand names, strip marketing
    language, etc.).

    \item \textbf{Transformation:}
    The debiasing model rewrites
    \[
      y^{\text{w/ad}}(x) 
      \;\longmapsto\; \hat{y}^{\text{w/o ad}}(x),
    \]
    filtering out the promotional material while preserving content.
\end{enumerate}
This method is efficient because it only requires a single inference call, but its success
depends on how accurately the secondary model detects and removes ad-specific tokens.
In particular, since commercial recommendations could appear indistinguishable from actual advice, it is unclear whether these methods are likely to be successful.

\subsection{Approach 2: Multi-sampling and aggregation}

A second approach is to generate multiple ad-influenced outputs for the same
query under varied serving conditions, then combine them to extract the common,
non-promotional core.
This method follows the intuition that changing platforms or other settings may lead to differing profiles of the same user, leading to shifts in the distribution of targeted advertisements.

\begin{enumerate}
    \item \textbf{Multi-sampling:}
    We obtain \(\smash{y^{\text{w/ad}}_1(x)} \dots
    y^{\text{w/ad}}_k(x)\)\,, each generated under slightly different parameters
    (e.g., user cookies, geolocation, synthetic personas) in order to elicit
    different sets of ads.

    \item \textbf{Aggregation:}
    By comparing these \(k\) outputs, we detect the stable text segments shared
    across all samples, filtering out brand- or sponsor-specific components that
    appear only in certain ones. Formally,
    \[
      \bigl\{\,y^{\text{w/ad}}_i(x)\bigr\}_{i=1}^k
      \;\longmapsto\; \hat{y}^{\text{w/o ad}}(x).
    \]
    A smaller, ad-free LLM (or comparable filtering logic) can thus approximate the
    original non-promotional output by discarding variable commercial elements.
\end{enumerate}
There are also recent works that propose statistical tests for identifying differences in the distributions of language model outputs~\citep{NEURIPS2024_e01519b4,modarressi2025causal}.
These types of methods provide a rigorous means for quantifying commercial biases in models.

\paragraph{Limitations and discussion}
Both approaches rely on the availability of an ad-free model.
Direct debiasing (Approach~1) offers a succinct, single-step pipeline, but it often presupposes knowledge
of the types of ads or sponsor references to remove. Multi-sampling and
aggregation (Approach~2) leverages variability in the way ads are served across
different contexts, but it requires multiple queries to the large LLMs.
Furthermore, this approach assumes that these contextual changes will indeed yield distinct promotional content.
Together, these strategies provide orthogonal means for eliminating
commercial biases from an LLM's output, thereby preserving user agency. 







\section{Open questions and future directions in generative advertising}
\label{sec:open-questions}

Although generative advertising brings new opportunities for monetization, it also poses several open questions about norms, regulations, and technical safeguards.
Below, we briefly outline key areas for future investigation.

\paragraph{Dynamic advertising and regulatory compliance}
Unlike static advertisements with well-defined formats, generative AI outputs are shaped by individual queries and stochastic model behavior~\citep{wei2022chain}. 
Enforcing existing disclosure requirements~\citep{Evans2017DisclosingII} becomes more challenging when sponsored text is woven into real-time user interactions.
It is not yet clear how regulators will verify compliance when model outputs differ across users or sessions~\citep{wangchain}.
Techniques for reproducibility~\citep{staudinger2024reproducibility}, logging, and automated auditing~\citep{rastogi2023supporting} must be adapted to accommodate this inherent variability.

\paragraph{User autonomy and consent}
Many current regulations require that users be given an option to opt out of \emph{targeted} advertising~\citep{goldfarb2011privacy}.
However, given that large language models can endorse sponsored content even in the absence of explicit personalization~\citep{johnson2020consumer}, more nuanced definitions of ``targeting'' may be required.
From the platform's standpoint, there is also an open question of how to enable individuals to limit commercial interactions, without entirely eliminating the commercial viability of advertising-based business models.

\paragraph{Transparency versus user experience}
Studies show that users implicitly trust outputs from generative AI, which are viewed as objective or data-driven~\citep{skjuve2024people}.
This trust may be eroded over time if sponsored content is improperly disclosed or covertly embedded~\citep{carlson2015news}.
However, disclosing ads too prominently can also detract from fluid interactions and decrease user satisfaction~\citep{kim2019seeing}.
Finding the right balance between transparency and user experience is an ongoing challenge.

\paragraph{Longitudinal impact on model alignment}
Section~\ref{sec:system} highlighted that user feedback loops drive the continual refinement of large language models~\citep{ouyang2022training,wang2023aligning}.
Notably, introducing ads as part of the training data risks blending commercial incentives with factual or unbiased knowledge~\citep{Feizi2023OnlineAW}.
Determining how to incorporate sponsored outputs into alignment pipelines without systematically biasing the model remains an unsolved problem.
There is also a need for practical strategies to surgically remove sponsored content from a trained model.

\paragraph{Evaluation metrics and debiasing strategies}
It is not yet clear what metrics best reflect commercial bias in LLM outputs~\citep{guo2022survey}.
While bias detection and factuality have established evaluation practices~\citep{limisiewicz2023debiasing,weilong}, quantifying \emph{commercial} biases requires new methodologies and datasets, based on existing advertising practices~\citep{zhu2025independence}.
Furthermore, while Section~\ref{sec:bias} presented strategies for mitigating sponsored content, many open questions remain regarding the implementation, scope, and reliability of these methods.
\section{Conclusion}
The increasing prominence of AI systems opens the door to generative advertising -- dynamic and personalized AI responses, which influenced by commercial factors.
As a community, we must ensure that AI systems remain safe and transparent.
Current technologies and policies only address a fraction of the open questions, whose resolution will require coordinated efforts across the machine learning community, as well as collaborations with the social sciences and regulatory agencies.
In exploring these frontiers,
researchers and model providers have the opportunity to ensure that the interactions between AI platforms, advertisers, and consumers are rooted in trust and user autonomy.


\newpage
\bibliographystyle{icml2025}
\bibliography{references}

\appendix

\section{Additional figures}

\begin{figure}[ht]
\centering
\includegraphics[width=\linewidth]{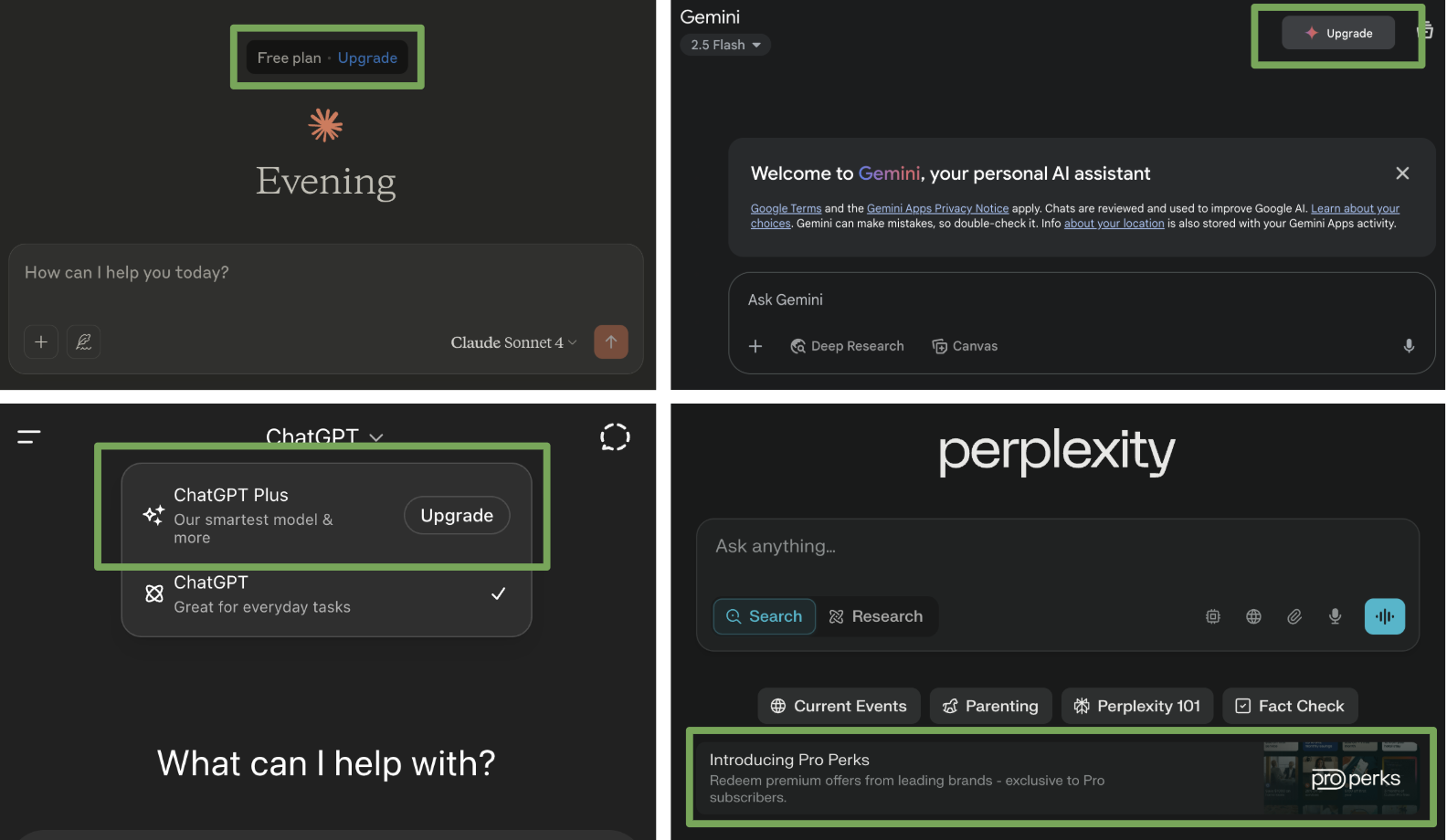}
\caption{Common AI systems prominently display static advertisements for premium features (green).
Screenshots were captured in May 2025.}
\label{fig:ad}
\end{figure}

\end{document}